\documentclass[10pt,fleqn]{article} 
\usepackage[preprint]{tmlr}

\usepackage{amsmath,amsfonts,bm}

\newcommand{\br}[1]{\left(#1\right)}
\newcommand{\av}[1]{\left\langle{#1}\right\rangle}
\newcommand{\ndist}[3]{{\cal{N}}\br{#1 \mid #2,#3}}
\newcommand{\norm}[1]{\left\Vert #1 \right\Vert}

\def\eqref#1{equation(\ref{#1})}
\def\obref#1{objective(\ref{#1})}

\def\1{\bm{1}}

\DeclareMathAlphabet{\mathsfit}{\encodingdefault}{\sfdefault}{m}{sl}
\SetMathAlphabet{\mathsfit}{bold}{\encodingdefault}{\sfdefault}{bx}{n}

\def\gF{{\mathcal{F}}}

\def\gU{{\mathcal{U}}}

\newcommand{\Ls}{\mathcal{L}}
\newcommand{\R}{\mathbb{R}}

\newcommand{\sq}[1]{\left[ {#1} \right]}

\usepackage{mathtools}

\newcommand{\shortoverleftarrow}[1]{%
  \overset{\mathclap{\raisebox{-1.5pt}{\scalebox{0.7}{$\leftarrow$}}}}{#1}%
}

\newcommand{\shortoverrightarrow}[1]{%
  \overset{\mathclap{\raisebox{-1.5pt}{\scalebox{0.7}{$\rightarrow$}}}}{#1}%
}

\newcommand{\fmu}{\shortoverrightarrow{\mu}}
\newcommand{\fSigma}{\shortoverrightarrow{\Sigma}}
\newcommand{\bmu}{\shortoverleftarrow{\mu}}

\usepackage{hyperref}
\usepackage{url}
\usepackage{amsthm}      
\usepackage{thmtools}    
\usepackage{booktabs}
\usepackage{float}       

\usepackage{thm-restate} 
\usepackage{graphicx}
\usepackage{enumitem}

\declaretheorem[name=Lemma]{lemma}

\usepackage{xcolor}

\title{A Discrete-Time Variational Objective for Degenerate Diffusion and Equivalence to Flow Matching}

\author{\name Rares Dolga \email rares.dolga.16@ucl.ac.uk \\
      \addr Foundational AI CDT,\\
      University College London
      \AND
      \name Harshad Dahake \email harshad.dahake.25@ucl.ac.uk \\
      \addr Centre for Artificial Intelligence, \\ University College London
      \AND
      \name David Barber \email david.barber@ucl.ac.uk\\
      \addr Centre for Artificial Intelligence, \\ University College London, \\
      UiPath}

\def\month{MM}  
\def\year{YYYY} 
\def\openreview{\url{https://openreview.net/forum?id=XXXX}} 

\begin{document}

\maketitle

\begin{abstract}
Diffusion models and more recently Flow models have widespread application. In this work we consider degenerate Diffusion models in which the per-step generator is deterministic. We construct a modified variational framework and show a formal equivalence between Flow models and Degenerate Diffusion models. For our discrete-time variational objective we show that the optimal degenerate generator recovers the data distribution in the limit of an infinite number of generator steps. Our work therefore  provides a unified perspective, showing that discrete-time implementations of flow matching and the discrete-time Degenerate Diffusion model can be considered within the same simple variational framework, removing the need for the continuous-time machinery normally used to justify flow matching models.
\end{abstract}



\section{Introduction}
Diffusion models are a widely used class of generative models~\citep{sohl2015deep, ho2020ddpm, song2019generative, song2021scoresde}, with notable success in domains such as image and video generation~\citep{dhariwal2021diffusion, rombach2022ldm, ho2022video, blattmann2023svd}. A fixed forward process corrupts the data with Gaussian noise of increasing scale, until it is indistinguishable from pure noise, and a stochastic generator is trained to undo it one step at a time.
In their variational derivation~\citep{sohl2015deep, ho2020ddpm, kingma2021vdm}, the training objective is a bound that decomposes into a sum of per-step KL divergences between the forward process and the generator.
Deterministic generators, however, have been found to give better samples in fewer steps~\citep{song2021ddim, karras2022edm}.
A deterministic generator is obtained by making each of its steps a Dirac delta function, in which case the KL divergence at each step is ill-defined.
DDIM~\citep{song2021ddim} bypasses this issue by training a stochastic model and setting its variance to zero only during sampling. However, this introduces a mismatch between the training objective, which assumes a stochastic per-step generator, and the use of a deterministic per-step generator after training.

Flow models~\citep{lipman2023flowmatchinggenerativemodeling, liu2023rectifiedflow, albergo2023building} are another important class of deterministic generators, now used at scale~\citep{esser2024sd3}. Their justification is based on continuous-time ODEs and the properties of flows, and, like the probability-flow ODE of score-based models~\citep{song2021scoresde}, they are derived outside the variational framework altogether.

A variational perspective on deterministic generators offers three benefits. First, it provides a training objective for the generator that is actually used for sampling: the per-step generator is deterministic both during and after training, so there is no mismatch of the kind introduced by DDIM. Second, it works directly with the discrete-time chain that is implemented in practice, rather than with the continuous-time ODE on which the justification of flow models rests. Third, it explains the freedom in weighting the per-step loss terms. The heuristic weightings used in practice, such as that of EDM~\citep{karras2022edm}, arise as different choices of a single free parameter of the objective (Section~\ref{sec:correctness}).


Since the KL divergence is ill-defined for a deterministic generator, we use the spread divergence~\citep{spreaddivergence}, which smooths both distributions with the same Gaussian noise so that the divergence between them is well-defined. We then define a discrete-time spread variational objective and show that, for a finite training set, the optimal deterministic generator recovers the data distribution as the number of steps tends to infinity (Theorem~\ref{thm:marginals}). Furthermore, the objective is equivalent to a discrete-time implementation of flow matching for a specific choice of spread noise and the linear schedule (Section~\ref{sec:flow-equiv}). This therefore also justifies why flow matching converges to the data distribution as the number of steps tends to infinity, without requiring the continuous-time machinery typically used to justify flow models.

\section{Background}
\subsection{Diffusion \label{sec:bg-diffusion}}
A standard diffusion model is defined by a forward noising process $q(x_{1:T} \mid x_0) = \prod_{t=1}^T q(x_t \mid x_{t-1},x_0)$, where $x_{1:T} = (x_1,\ldots,x_T)$, each forward probability $q(x_t \mid x_{t-1},x_0)$ is linear-Gaussian, and $q(x_0)$ is the data distribution on $\R^d$. In practice, $q(x_0)$ is the empirical distribution of a finite training set, $q(x_0) = \frac{1}{N}\sum_{n=1}^N \delta(x_0 \mid x_0^n)$ where  $\delta(x \mid y)$ is a Dirac delta for $x$ with all mass placed at $y$. This is paired with a generative model $p_\theta(x_{0:T}) = p(x_T)\prod_{t=1}^T p_\theta(x_{t-1} \mid x_t)$ where $p_\theta(x_{t-1} \mid x_t) = \ndist{x_{t-1}}{\mu_\theta(x_t)}{\beta_tI}$ with learnable mean $\mu_\theta$ and fixed variance $\beta_t > 0$. As is standard, the network $\mu_\theta$ also receives the timestep $t$, which we suppress in the notation throughout.
The forward process is defined such that $q(x_T \mid x_0) \approx p(x_T)=\ndist{x_T}{0}{I}$.
Training minimises an upper bound on the negative log-likelihood, which follows from $\mathrm{KL}[q(x_0) \,\|\, p_\theta(x_0)] \le \mathrm{KL}[q(x_{0:T}) \,\|\, p_\theta(x_{0:T})]$:
\begin{multline}
  -\av{\log  p_\theta(x_0)}_{q(x_0)}
    \le \sum_{t=2}^T
        \av{\mathrm{KL}[q( x_{t-1} \mid x_t, x_0) \,\|\, p_\theta(x_{t-1} \mid x_t) ]}_{q(x_t \mid x_0) q(x_0)} \\
    - \av{\log p_\theta(x_0 \mid x_1)}_{q(x_1 \mid x_0)q(x_0)}
    + C
\label{eq:diffu-loss}
\end{multline}
where $\av{\:\cdot\:}_p$ represents expectation with respect to the distribution $p$ and the constant $C = \av{\mathrm{KL}[q(x_T \mid x_0) \,\|\, p(x_T)]}_{q(x_0)}$ is independent of the model parameters (Appendix~\ref{app:diffu}). Setting $q(x_{t-1} \mid x_t, x_0) = \delta(\cdot \mid x_0)$ at $t=1$, the right-hand side of \eqref{eq:diffu-loss} is, up to additive constants\footnote{The two differ by $C$ and, at $t=1$, by the entropy of a delta, which is not finite, so this identity is formal. It becomes exact once the spread divergence is applied (Section~\ref{sec:spread-obj}).} that do not depend on $\theta$, equivalent to
\begin{equation}
  L(\theta) = \sum_{t=1}^T
    \av{\mathrm{KL}[q(x_{t-1} \mid x_t, x_0) \,\|\, p_\theta(x_{t-1} \mid x_t)]}_{q(x_t \mid x_0) q(x_0)}.
  \label{eq:diffu-loss-unified}
\end{equation}

\subsection{Spread Divergence \label{sec:bg-spread}}
A divergence $\mathrm{Div}(p \,\|\, q)$ between two distributions with different supports, or without densities, may be ill-defined.
The spread divergence~\citep{spreaddivergence} resolves this by convolving both distributions with the same noise kernel $k(y \mid x)$. We use the Gaussian kernel $k(y \mid x) = \ndist{y}{x}{\sigma^2I}$ with $\sigma > 0$:
\begin{align}
  \tilde{p}(y) = \int dx \, k(y \mid x)\, p(x),
  \qquad
  \tilde{q}(y) = \int dx \, k(y \mid x)\, q(x).
  \label{eq:spread-def}
\end{align}
The spread distributions $\tilde{p}$ and $\tilde{q}$ have the same support for any $p$ and $q$, so $\mathrm{Div}(\tilde{p} \,\|\, \tilde{q})$ is well-defined.
Since the Gaussian kernel depends only on $y - x$, the spread operation is a convolution, $\tilde{p} = k * p$. Under the Fourier transform $\mathcal{F}$, this becomes a product, $\mathcal{F}\{\tilde{p}\} = \mathcal{F}\{k\}\,\mathcal{F}\{p\}$.
Moreover, since $\mathcal{F}\{k\}$ is itself a Gaussian and nowhere zero, it can be divided out, and each of the following implications holds in both directions:
\begin{equation}
  \mathrm{Div}(\tilde{p} \,\|\, \tilde{q}) = 0
  \;\Leftrightarrow\; \tilde{p} = \tilde{q}
  \;\Leftrightarrow\; \mathcal{F}\{k\}\,\mathcal{F}\{p\} = \mathcal{F}\{k\}\,\mathcal{F}\{q\}
  \;\Leftrightarrow\; \mathcal{F}\{p\} = \mathcal{F}\{q\}
  \;\Leftrightarrow\; p = q.
\end{equation}
Hence the spread divergence is zero if and only if $p = q$.

\section{Degenerate Diffusion Model \label{sec:model-def}}
We define the Degenerate Diffusion Model, in which each step of both the forward noising process and the generator is deterministic. In the forward process, the only randomness comes from the data point $x_0$ and a single Gaussian noise sample $\epsilon$ shared by all timesteps; in the generator, it comes only from the initial state $x_T$.
We define a strictly increasing schedule $\lambda_t$, $t \in \{0, 1, \ldots, T\}$, with $\lambda_0 = 0$ and $\lambda_T = 1$, and the degenerate forward distributions $q(x_t \mid x_0,\epsilon) = \delta(x_t \mid (1-\lambda_t)x_0 + \lambda_t \epsilon)$ with $\epsilon \in \R^d$ and $q(\epsilon) = \ndist{\epsilon}{0}{I}$, so that $q(x_T)=\ndist{x_T}{0}{I}$.
Hence, the forward noising process
\begin{align}
  q(x_{1:T} \mid x_0) &= \int d \epsilon \, q(\epsilon) \prod_{t=1}^T q(x_t \mid x_0,\epsilon) = q(x_1 \mid x_0) \prod_{t=2}^T q(x_t \mid x_{t-1}, x_0)
\end{align}
is a degenerate Gaussian on $\R^{Td}$ (its covariance has rank $d$)\footnote{The noise $\epsilon=\big(x_{t-1}-(1-\lambda_{t-1})x_0\big)/\lambda_{t-1}$ is recoverable from $(x_{t-1},x_0)$, so $x_t$ is a deterministic function and independent of $x_{1:t-2}$ given $(x_{t-1},x_0)$. Only $q(x_1\mid x_0)$ is stochastic; each $q(x_t\mid x_{t-1},x_0)$, $t\ge2$, is a delta.}.
We also define the generator model $p_\theta(x_{t-1} \mid x_t) = \delta(x_{t-1} \mid \mu_\theta(x_t))$ with $p(x_T) = \ndist{x_T}{0}{I}$.
Generating a sample means running the generator from noise, $\hat{x}_T \sim \ndist{\cdot}{0}{I}$ and $\hat{x}_{t-1} = \mu_\theta(\hat{x}_t)$ for $t = T, \ldots, 1$, with $\hat{x}_0$ the generated sample. We call this iteration the \emph{sampler}, and the resulting sequence $\hat{x}_T, \ldots, \hat{x}_0$ the \emph{sampled chain} (the hat distinguishes its states from those of the forward process).
Each term of the objective \eqref{eq:diffu-loss-unified} compares the generator $p_\theta(x_{t-1} \mid x_t)$ with the data-conditioned backward distribution $q(x_{t-1} \mid x_t, x_0)$. The following Lemma gives simple expressions for this distribution and for $q(x_t \mid x_0)$, and shows that, like the generator, $q(x_{t-1} \mid x_t, x_0)$ is a delta function.

\begin{restatable}{lemma}{xtgivenx}{}
  \label{th:xtgivenx}
  Let $(\lambda_t)_{t=0}^T$ be strictly increasing with $\lambda_0 = 0$, and let $x_1, \ldots, x_T$ be defined by
  \begin{equation}
    x_t \;=\; (1-\lambda_t) x_{0} + \lambda_t \, \epsilon,
    \qquad \epsilon \sim \ndist{\cdot}{0}{I}.
    \label{eq:forward-path}
  \end{equation}
  Then, for $t = 1, \ldots, T$,
  \begin{equation}
    q(x_t \mid x_0) \;=\; \ndist{x_t}{\fmu_t}{\fSigma_t}
   \quad \text{and} \quad q(x_{t-1} \mid x_t,x_0) = \delta\big(x_{t-1} \bigm|\bmu_{t-1}(x_t, x_0)\big),
  \end{equation}
 where
 \begin{equation}
     \fmu_t = (1 - \lambda_t)x_0,
    \quad \fSigma_t = \lambda_t^2 I, \quad \bmu_{t-1}(x_t, x_0)= \br{1- \frac{\lambda_{t-1}}{\lambda_t}} x_{0} + \frac{\lambda_{t-1}}{\lambda_t} x_t.
 \end{equation}
\end{restatable}

The proofs are given in Appendix~\ref{lemma1_proofs}.
By Lemma~\ref{th:xtgivenx} and the definition of the generator, both $q(x_{t-1} \mid x_t, x_0)$ and $p_\theta(x_{t-1} \mid x_t)$ in each term of \eqref{eq:diffu-loss-unified} are delta functions. For the Degenerate Diffusion Model, each KL divergence in the objective is therefore ill-defined. 

\subsection{Spread Divergence Objective \label{sec:spread-obj}}

To resolve the issue with KL divergence between degenerate distributions, we use the spread divergence.
This smooths the deterministic point masses by convolving them with noise distributions, thereby yielding a finite, well-behaved divergence. Applying the spread divergence to each term of \eqref{eq:diffu-loss-unified} yields our \emph{spread objective}:
\begin{align}
  L'(\theta) = \sum_{t=1}^T
    \av{ \mathrm{KL}[\tilde{q}(y_{t-1} \mid x_t, x_0) \,\|\, \tilde{p}_\theta(y_{t-1} \mid x_t)] }_{q(x_t \mid x_0) q(x_0)},
  \label{eq:spread-loss}
\end{align}
where the spread distributions $\tilde{q}$ and $\tilde{p}_\theta$ are obtained by convolving the delta distributions with a kernel $k(y_{t-1} \mid x_{t-1}) = \ndist{y_{t-1}}{x_{t-1}}{\sigma^2_{t}I}$. Throughout the paper we refer to $\sigma_t$ as the \emph{spread noise level} and to $\sigma_t^2$ as the \emph{spread variance}:
\begin{align}
  \tilde{q}(y_{t-1} \mid x_t, x_0)
    &= \int dx_{t-1} \, k(y_{t-1} \mid x_{t-1})\, \delta(x_{t-1} \mid \bmu_{t-1}(x_t, x_0))
     = \ndist{y_{t-1}}{\bmu_{t-1}(x_t, x_0)}{\sigma^2_tI} \\
  \tilde{p}_\theta(y_{t-1} \mid x_t)
    &= \int dx_{t-1} \, k(y_{t-1} \mid x_{t-1})\, \delta(x_{t-1} \mid \mu_\theta(x_t))
     = \ndist{y_{t-1}}{\mu_\theta(x_t)}{\sigma^2_tI}.
  \label{eq:spread-conditionals}
\end{align}
Both spread distributions are normal with covariance $\sigma^2_tI$, differing only in their means $\bmu_{t-1}(x_t, x_0)$ and $\mu_\theta(x_t)$. 
The spread divergence is therefore finite and the objective becomes:
\begin{align}
L'(\theta) = \sum_{t=1}^T \frac{1}{2\sigma^2_t}
    \av{ \norm{\bmu_{t-1}(x_t, x_0) - \mu_\theta(x_t)}^2 }_{q(x_t \mid x_0) q(x_0)}.
    \label{eq:spread-final-loss}
\end{align}

A natural alternative is to spread the joint distributions $q(x_{0:T})$ and $p_\theta(x_{0:T})$, adding Gaussian noise to every variable, $y_t = x_t + \sigma_t \eta_t$ with $\eta_t \sim \ndist{\eta_t}{0}{I}$. This also gives a valid divergence, but the noise now corrupts every clean $x_t$, including the inputs of the generator. Since the generator is deterministic given $x_T$, its clean trajectory is $x_t = \mu_\theta^{(T-t)}(x_T)$, where $\mu_\theta^{(n)}$ denotes $n$ successive applications of $\mu_\theta$, and the spread generator becomes
\begin{equation}
  \tilde{p}_\theta(y_{0:T}) = \int dx_T \, p(x_T) \prod_{t=0}^{T} \ndist{y_t}{\mu_\theta^{(T-t)}(x_T)}{\sigma_t^2 I}.
\end{equation}
Evaluating it would require integrating over $x_T$ and running the generator through all of its steps; the resulting objective is intractable and does not decompose over timesteps. In contrast, each term of our simpler  \obref{eq:spread-loss} conditions on a clean $x_t$, sampled directly from the forward process $q(x_t \mid x_0)$, and spreads only $x_{t-1}$; this gives the Gaussians \eqref{eq:spread-conditionals} and the per-step mean squared error \eqref{eq:spread-final-loss}.

\subsubsection{Objective Reparameterisation \label{sec:reparam}}
In \obref{eq:spread-final-loss}, the network directly predicts the next state $x_{t-1} = \mu_\theta(x_t)$. As in standard diffusion models, we can instead use alternative parameterisations.
%
%
For the velocity parameterisation, analogous to that of \mbox{\citet{salimans2022progressive}}, we define $v_t(x_0) = \frac{x_t - x_0}{\lambda_t}$, so that $x_0 = x_t - \lambda_t v_t(x_0)$.
Substituting $x_0$ in $\bmu_{t-1}(x_t, x_0)$ gives $\bmu_{t-1}(x_t, x_0) = x_t - \Delta\lambda_t\, v_t(x_0)$, where $\Delta\lambda_t = \lambda_t - \lambda_{t-1}$ is the step size. The velocity parameterisation is therefore $\mu_\theta(x_t) = x_t - \Delta\lambda_t\, v_\theta(x_t)$
and results in the equivalent reparameterised objective
\begin{align}
    L'(\theta) = \sum_{t=1}^T \frac{(\Delta\lambda_t)^2}{2\sigma^2_t}
    \av{ \norm{v_t(x_0) - v_\theta(x_t)}^2}_{q(x_t \mid x_0) q(x_0)}.
    \label{eq:spread-velocity-loss}
\end{align}

Using the definition of the noising process in \eqref{eq:forward-path}, the velocity can also be written in terms of the data and the noise alone, $v_t(x_0) = \epsilon - x_0$. For fixed $(x_0, \epsilon)$, the target velocity is therefore independent of $t$, since all $x_t$ lie on the straight line from the data point $x_0$ to the noise sample $\epsilon$.

The spread noise level enters \obref{eq:spread-velocity-loss} only through the weight $\omega_t = (\Delta\lambda_t)^2 / (2\sigma_t^2)$ of each mean squared error term. Assuming the network $v_\theta$ is flexible enough to represent any function of $(x_t, t)$, each term can be minimised separately and, as is standard for a mean squared error, its minimiser is the conditional mean $\av{v_t(x_0)}_{q(x_0 \mid x_t)}$~\citep{karras2022edm}. The minimiser therefore does not depend on $\sigma_t$. Conversely, since $\sigma_t$ is free, any fixed positive weights $\omega_t$ are obtained by the choice $\sigma_t^2 = (\Delta\lambda_t)^2 / (2\omega_t)$. A custom weighting of the mean squared error across timesteps, such as that of EDM~\citep{karras2022edm}, therefore corresponds to a spread divergence with a different spread noise level at each step. 
In Section~\ref{sec:correctness} we show that each such objective minimises a divergence to the backward distribution $q(x_{t-1} \mid x_t)$, which is what makes these weightings principled.

\subsection{What the Spread Objective Learns \label{sec:correctness}}
Each term of the spread objective \eqref{eq:spread-loss} compares the generator with the data-conditioned backward distribution $q(x_{t-1} \mid x_t, x_0)$, which depends on the data point $x_0$, whereas the generator only sees $x_t$. The distribution that the generator should match is instead the \emph{backward distribution}
\begin{equation}
  q(x_{t-1} \mid x_t) = \int dx_0 \, q(x_0 \mid x_t)\, q(x_{t-1} \mid x_t, x_0),
  \label{eq:backward}
\end{equation}
which does not depend on $x_0$. If $p_\theta(x_{t-1} \mid x_t) = q(x_{t-1} \mid x_t)$ for every $t$, then, since $p(x_T) = q(x_T)$, the marginals of the generator satisfy $p_\theta(x_{t-1}) = \int dx_t \, q(x_{t-1} \mid x_t)\, q(x_t) = q(x_{t-1})$ for every $t$ by induction, and in particular its samples are distributed as the data, $p_\theta(x_0) = q(x_0)$.
In this subsection we show that minimising the spread objective $L'$ is equivalent to minimising a spread divergence between the generator and the backward distribution, so that the objective targets exactly this distribution.

The spread of the backward distribution $q(x_{t-1} \mid x_t)$ is
\begin{align}
  \tilde{q}(y_{t-1} \mid x_t)
    = \int dx_{t-1}\, k(y_{t-1} \mid x_{t-1})\, q(x_{t-1} \mid x_t)
    = \int dx_0\, q(x_0 \mid x_t)\, \tilde{q}(y_{t-1} \mid x_t,x_0).
  \label{eq:spread-backward}
\end{align}
Then, the spread divergence between the backward distribution and the generator becomes:
\begin{align}
  \mathcal{D}(\theta) = \sum_{t=1}^T
    \av{\mathrm{KL}[\tilde{q}(y_{t-1} \mid x_t) \,\|\, \tilde{p}_\theta(y_{t-1} \mid x_t)]}_{q(x_t)},
  \label{eq:natural-spread-obj}
\end{align}
which, following the properties established in Section~\ref{sec:bg-spread}, is zero if and only if $p_\theta(x_{t-1} \mid x_t) = q(x_{t-1} \mid x_t)$
for every $t$ and $x_t$.

\begin{restatable}[Decomposition of the spread objective]{lemma}{spreaddecomp}
  \label{th:spread-decomp}
  For every $\theta$,
  \begin{align}
    L'(\theta) = \mathcal{D}(\theta) + C,
  \end{align}
  where $C$ is a constant independent of $\theta$.
\end{restatable}



The proof is given in Appendix~\ref{app:proofs}.
The divergence $\mathcal{D}(\theta)$ cannot be evaluated directly, since the spread backward distribution \eqref{eq:spread-backward} requires the posterior $q(x_0 \mid x_t)$, whereas $L'$ requires only samples of $(x_0, x_t)$. Since $C$ does not depend on $\theta$, the two have the same minimiser, so training with the tractable objective $L'$ targets the backward distribution.

Lemma~\ref{th:spread-decomp} holds for any $\sigma_t > 0$. Since each choice of $\sigma_t$ corresponds to a fixed positive weighting of the per-step mean squared errors (Section~\ref{sec:reparam}), every such weighting, including the heuristic weightings used in practice such as that of EDM~\citep{karras2022edm}, minimises a spread divergence to the backward distribution. Within our framework these weightings are therefore principled: they correspond to different choices of the spread and all share the same minimiser.

\subsection{Convergence to the Data Distribution}
\label{sec:convergence}

Since the sampler is deterministic given $\hat{x}_T$, it cannot in general match the backward distribution $q(x_{t-1} \mid x_t)$ (Section~\ref{sec:correctness}) which is a mixture distribution over datapoints. We now show that, at the optimum of the spread objective,  generator samples $\hat{x}_0$ nevertheless converge to the data distribution as $T \to \infty$.

We consider a finite training set of $N \ge 2$ distinct points, $q(x_0) = \frac{1}{N}\sum_{n=1}^N \delta(x_0 \mid x_0^n)$  as defined in Section~\ref{sec:bg-diffusion}. The forward marginals are then mixtures of Gaussians,
\begin{equation}
  q_t(x_t) := q(x_t) = \frac{1}{N}\sum_{n=1}^N \ndist{x_t}{(1-\lambda_t)x_0^n}{\lambda_t^2 I},
\end{equation}
and the posterior over training points is
\begin{equation}
  q(x_0^n \mid x_t) = w_n^t(x_t), \qquad w_n^t(x) = \frac{\exp\{-\norm{x - (1-\lambda_t)x_0^n}^2/(2\lambda_t^2)\}}{\sum_{k=1}^N \exp\{-\norm{x - (1-\lambda_t)x_0^k}^2/(2\lambda_t^2)\}}.
\end{equation}
Substituting this into \eqref{eq:backward} and using Lemma~\ref{th:xtgivenx}, the backward distribution is a mixture of deltas, one per training point:
\begin{equation}
  q(x_{t-1} \mid x_t) = \sum_{n=1}^N w_n^t(x_t)\, \delta\big(x_{t-1} \bigm| \bmu_{t-1}(x_t, x_0^n)\big).
  \label{eq:backward-mixture}
\end{equation}
At the optimum of the spread objective, the generator outputs the mean of this mixture. Each term of \eqref{eq:spread-final-loss} is a mean squared error in $\mu_\theta(x_t)$, so its minimiser is
\begin{equation}
  \mu_t^*(x) = \sum_{n=1}^N w_n^t(x)\, \bmu_{t-1}(x, x_0^n) = \bmu_{t-1}\big(x, m_t(x)\big), \qquad m_t(x) = \sum_{n=1}^N w_n^t(x)\, x_0^n,
\end{equation}
where the second equality holds because $\bmu_{t-1}$ is linear in $x_0$, and $m_t(x)$ is the posterior mean of the data. This optimum does not depend on how $\mu_\theta$ is parameterised. For the velocity parameterisation of Section~\ref{sec:reparam}, it corresponds to $v_t^*(x) = (x - m_t(x))/\lambda_t$. Replacing the mixture \eqref{eq:backward-mixture} by a delta at its mean introduces an error at every step, but we show that its accumulated effect vanishes as $T \to \infty$, so that the sampler converges weakly to the data distribution.


\begin{restatable}[Weak convergence]{theorem}{marginalsequal}\label{thm:marginals}
Let $q(x_0)$ be the empirical distribution of $N \ge 2$ distinct points $x_0^1, \ldots, x_0^N \in \R^d$. For each $T$, let the interpolation schedule be as in Section~\ref{sec:model-def}, and assume that the largest step size vanishes, $\max_t \Delta\lambda_t \to 0$ as $T \to \infty$. Let $\hat{x}_T, \ldots, \hat{x}_0$ be the sampled chain, and assume that the generator attains the exact optimum of the spread objective, $\mu_\theta(x) = \mu_t^*(x)$ for every $t$ and every $x \in \R^d$. Then there is a radius $\gamma_T > 0$, depending only on the schedule and the dataset, with $\gamma_T \to 0$ as $T \to \infty$, such that for every data point $x_0^m$,
\begin{align}
 \lim_{T \to \infty }P\br{\norm{\hat{x}_0 - x_0^m} \le \gamma_T} = \frac{1}{N}.
\end{align}
Consequently, the distribution of the generated sample $\hat{x}_0$ converges weakly to $q(x_0)$.
\end{restatable}

The theorem says that, starting from Gaussian noise, the generated sample $\hat{x}_0$ ends inside the ball of radius $\gamma_T$ around the training point $x_0^m$ with probability tending to $\frac{1}{N}$, for every $m$. Summing over the $N$ training points, $\hat{x}_0$ ends inside the ball around some training point with probability tending to one.
The radius $\gamma_T$ vanishes as the number of steps grows, so in the limit the generated samples are the training points themselves, each with probability $\frac{1}{N}$, which is exactly $q(x_0)$.

\paragraph{Proof sketch.}
We split the trajectory into two regimes, high and low noise levels.
At high noise, we compare the sampled chain with a reference chain that starts from the same noise but where, instead of averaging over the posterior $q(x_0 \mid x_t)$ at each step, the reference chain draws one training point from it.
This chain samples from the backward distribution, so its marginals are exactly $q_t$.
The two chains differ at each step by a zero-mean term with variance of order $(\Delta\lambda_t)^2$, so their expected squared distance at the end of this regime is upper bounded by an expression of the order of the largest step size which vanishes as $T \to \infty$ (Lemma~\ref{lem:phase1}).

At low noise, the posterior concentrates on a single training point.
Each data point $x_0^m$ gives one Gaussian component of $q_t$, centred at $(1-\lambda_t)x_0^m$, and this centre moves to $x_0^m$ as $\lambda_t \to 0$ (i.e., as $t \to 0$).
Once the sampled chain is close to such a centre, it stays close to it at every later step, and it ends very close to $x_0^m$.
The remaining distance vanishes exponentially fast as the last step size $\lambda_1$ goes to zero (Lemma~\ref{lem:trapping}).
Since the reference chain has marginals $q_t$, it is near each centre with probability close to $\frac{1}{N}$ at the step $\tau$ where the two regimes meet. By the first part of the argument, the sampled chain is close to the reference chain at step $\tau$, so it is also near each centre with probability close to $\frac{1}{N}$, and then stays near it until the end.
The full proof is in Appendix~\ref{app:convergence}.

\paragraph{Remarks.}
The theorem holds for any choice of the spread noise levels $\sigma_t$. They only weight the terms of the loss and do not change its optimum $\mu_t^*$, which is what the theorem uses (Section~\ref{sec:reparam}) for sampling.
The result assumes the exact optimum on a finite dataset, for which the sampler reproduces the training points.
It is therefore a consistency result for the objective and does not describe a trained network's generalisation.
Finally, the proof works purely in discrete-time, without relying on any continuous-time methods.

\section{Connection to Flow Matching}
\label{sec:flow-equiv}
While the forward path $x_t = (1-\lambda_t)x_0 + \lambda_t\epsilon$ of Section~\ref{sec:model-def} is the interpolant used by flow matching, we derived our objective and sampler solely from the variational bound. We now show that our objective, its minimiser and the resulting sampler are equivalent to those of flow matching.

We consider the standard conditional flow matching loss~\citep{lipman2023flowmatchinggenerativemodeling}
\begin{align}
    \Ls_{\mathrm{CFM}}(\theta) =
    \av{ \norm{ u_s(x_s \mid x_0) - u_\theta(x_s, s) }^2 }_{\gU(s)\, q(x_s \mid x_0)\, q(x_0)},
    \label{eq:fm-loss1}
\end{align}
where $\gU(s)$ is the uniform distribution on $[0,1]$, $u_\theta$ is the learned velocity field, and $u_s(x_s \mid x_0) = \frac{dx_s}{ds}$ is the conditional velocity field. Variants that sample $s$ from a non-uniform density instead of $\gU(s)$~\citep{esser2024sd3} correspond to a weighting of this loss over $s$, which our framework accommodates as a choice of spread (Section~\ref{sec:reparam}).

Let $\lambda : [0,1] \to [0,1]$ be twice continuously differentiable with $\lambda'(s) > 0$, $\lambda(0) = 0$ and $\lambda(1) = 1$, and let $x_s = (1-\lambda(s))x_0 + \lambda(s)\epsilon$ for $s \in [0,1]$, so that $s = 0$ corresponds to the data and $s = 1$ to the noise
\footnote{This is the reverse of the time convention of \citet{lipman2023flowmatchinggenerativemodeling}, who place the data at time $1$.}.
Differentiating $x_s$ gives
\begin{align}
    u_s(x_s \mid x_0) = \frac{d(1-\lambda(s))}{ds}x_0 + \frac{d\lambda(s)}{ds}\epsilon
    = \lambda'(s)(\epsilon - x_0)
    = \lambda'(s)\, v_s(x_0),
\end{align}
with $v_s(x_0) = \epsilon - x_0$ as in Section~\ref{sec:reparam}\footnote{We keep the subscript in $v_s(x_0)$, although for fixed $(x_0, \epsilon)$
it does not depend on $s$, for easy comparison with the previous sections.}.
We reparameterise the flow network by the velocity network of Section~\ref{sec:reparam},
$u_\theta(x_s, s) := \lambda'(s)\, v_\theta(x_s, s)$, where we now make its time conditioning explicit. The loss \eqref{eq:fm-loss1} then becomes
\begin{align}
    \Ls_{\mathrm{CFM}}(\theta) = \int_0^1 \lambda'(s)^2\, \underbrace{\av{\norm{v_s(x_0) - v_\theta(x_s, s)}^2}_{q(x_s \mid x_0)q(x_0)}}_{=:\, \ell(s)}\, ds.
    \label{eq:fm-loss2}
\end{align}

To relate \eqref{eq:spread-velocity-loss} to \eqref{eq:fm-loss2}, we place our
discrete steps on the time axis, $s_t = t/T$ and $\lambda_t = \lambda(s_t)$, and evaluate
the network at the grid times, $v_\theta(x_t) = v_\theta(x_t, s_t)$.
We first relate the step sizes $\Delta\lambda_t$ to $\lambda'$. By Taylor's theorem, for every $s$ and $h > 0$ with $s - h \in [0,1]$, $\lambda(s - h) = \lambda(s) - h\,\lambda'(s) + O(h^2)$. Since $\lambda''$ is continuous, and hence bounded on $[0,1]$, the constant in $O(h^2)$ is the same for every $s$.
We apply Taylor's theorem with $s = s_t$ and $h = 1/T$, so that $s - h = (t-1)/T = s_{t-1}$. Rearranging gives
\begin{equation}
  \Delta\lambda_t = \lambda(s_t) - \lambda(s_{t-1}) = \frac{\lambda'(s_t)}{T} + O(T^{-2}),
  \label{eq:step-size}
\end{equation}
where $O(T^{-2})$ is constant with respect to $t$. 

We now compare the two methods across three aspects: objective, minimiser, and sampler. The comparison holds for any schedule $\lambda$ as above; for the flow matching default $\lambda(s) = s$, $\Delta\lambda_t = 1/T$ exactly and all the $O(\cdot)$ terms below vanish.

\paragraph{Objective.}
Since $v_t(x_0) = \epsilon - x_0 = v_{s_t}(x_0)$, taking a constant spread noise level $\sigma_t = \sigma$ in \eqref{eq:spread-velocity-loss} gives
\begin{equation}
  2\sigma^2 T\, L'(\theta)
  = T \sum_{t=1}^{T} (\Delta\lambda_t)^2\, \ell(s_t)
  = \underbrace{\frac{1}{T}\sum_{t=1}^{T} \lambda'(s_t)^2\, \ell(s_t)}_{\text{Riemann sum}}
  + O(T^{-1}).
\end{equation}
Here we assume that $v_\theta$ is continuous in $(x, s)$ and grows at most polynomially in $x$, uniformly in $s$, that is $\norm{v_\theta(x, s)} \le A + B\norm{x}^k$ (for some integer $k \ge 1$, all $x$ and all $s \in [0,1]$), and that all moments of $q(x_0)$ are finite. The first is a mild regularity condition (analyses of flow matching and of probability-flow ODEs make similar assumptions on the learned network, such as differentiability or Lipschitz continuity in $x$~\citep{benton2024error, chen2023probabilityflow}). The second holds for any finite dataset. Under these assumptions $\ell$ is continuous on $[0,1]$, and hence bounded (Lemma~\ref{th:ell-continuous}).
The second equality follows from squaring \eqref{eq:step-size}, which gives $(\Delta\lambda_t)^2 = \lambda'(s_t)^2/T^2 + O(T^{-3})$, with the same constant for every $t$ because $\lambda'$ is continuous, and hence bounded, on $[0,1]$. Since $\ell$ is also bounded, multiplying by $\ell(s_t)$ keeps each error term $O(T^{-3})$. Summing the $T$ terms and multiplying by $T$ then gives $O(T^{-1})$.
Since $\ell$ is continuous, the Riemann sum converges to $\int_0^1 \lambda'(s)^2 \ell(s)\, ds$, so $2\sigma^2 T\, L'(\theta) \to \Ls_{\mathrm{CFM}}(\theta)$ as $T \to \infty$. Choosing $\sigma^2 = \frac{1}{2T}$\footnote{which satisfies the positive weight requirement} removes the prefactor, so that $L'(\theta) \to \Ls_{\mathrm{CFM}}(\theta)$ directly.

\paragraph{Minimiser.}
For any $T$, the minimiser of $L'$ at each grid time $s_t$ is $v^*(\cdot, s_t)$, where $v^*(x, s) = \av{\epsilon - x_0}_{q(x_0, \epsilon \mid x_s = x)}$ is the minimiser of $\Ls_{\mathrm{CFM}}$ (taken continuous in $s$).
The two objectives therefore yield the same velocity field at the grid times.

\paragraph{Sampler.}
The generator mean $\mu_\theta(x_t) = x_t - \Delta\lambda_t\, v_\theta(x_t, s_t)$ (Section~\ref{sec:reparam}) is exactly an Euler step of the ODE $dx/d\lambda = v_\theta$ in the variable $\lambda$.
Substituting first \eqref{eq:step-size} and then $u_\theta(x, s) = \lambda'(s)\, v_\theta(x, s)$ gives
\begin{align}
  \mu_\theta(x_t)
  &= x_t - \frac{1}{T}\, \lambda'(s_t)\, v_\theta(x_t, s_t) + O(T^{-2}) \nonumber \\
  &= x_t - \frac{1}{T}\, u_\theta(x_t, s_t) + O(T^{-2}).
\end{align}
That is, an Euler step of size $1/T$ of the flow matching ODE $dx/ds = u_\theta$, up to a correction that is small and vanishes as $T \to \infty$.

Hence, for $\sigma^2 = \frac{1}{2T}$, our objective converges to the flow matching loss with the same minimiser. Consequently, the theoretical results of Sections~\ref{sec:correctness} and~\ref{sec:convergence} apply directly to the flow matching loss and its weightings. For the default schedule $\lambda(s) = s$, the correspondence is exact at every $T$: our objective equals the flow matching loss evaluated on the uniform grid, $L'(\theta) = \frac{1}{T}\sum_{t=1}^T \ell(s_t)$, and our sampler is exactly its Euler sampler.

\section{Related Work}
DDIM~\citep{song2021ddim} with zero variance is the diffusion model closest to ours. Like our method, it samples with a deterministic generator, and its forward process reuses a single noise sample across all timesteps, though with a different noise schedule. The key difference is how the deterministic generator is obtained. DDIM trains a stochastic per-step generator and sets the variance to zero only at sampling time, so the generator assumed by the training objective is not the one used to produce samples. We instead apply spread divergence~\citep{spreaddivergence} to each step of the variational bound, which gives a well-defined training objective for the deterministic generator itself. As a result, the same generator is used for both training and sampling.

Flow matching~\citep{lipman2023flowmatchinggenerativemodeling, liu2023rectifiedflow} uses exactly our forward path, the straight line between the data point and the noise sample, but trains the generator with a regression loss rather than a variational objective. \citet{gao2025diffusionflow} show that, for such paths, flow matching and DDIM use the same sampler, and that their training losses differ only in how the network is parameterised and how the timesteps are weighted. Our model recovers this loss and sampler from a single variational objective (Section~\ref{sec:flow-equiv}), and the difference in weighting corresponds to a different choice of spread (Section~\ref{sec:reparam}).

We also show that the optimum does not depend on the spread noise level, which only sets the weight $\omega_t$ of each per-step loss term (Section~\ref{sec:reparam}). This relates to \citet{kingma2023understanding}, who show that weighted diffusion losses can be read as an ELBO of noise-perturbed data, provided that the forward process and the generator are Gaussian and the weighting is monotonic. In our framework, any fixed positive weighting $\omega_t$, monotonic or not, corresponds to a spread divergence.

Variational Flow Matching~\citep{eijkelboom2024vfm} also gives flow matching a variational interpretation, but in continuous time and of a different kind: the velocity field is parameterised through a variational approximation of the posterior $q(x_0 \mid x_t)$, while the generator remains the deterministic ODE. Our objective is instead a variational objective for the deterministic discrete-time generator itself.

Under the variational view, we also prove convergence of the deterministic sampler to the data distribution. Convergence has already been shown in the continuous-time setting of flow matching~\citep{benton2024error} and probability-flow ODEs~\citep{chen2023probabilityflow, li2023faster, gao2024wasserstein}.
Our proof instead works directly in discrete time, without the continuous-time arguments on which these results rely.
It holds at the exact optimum on a finite training set, where the sampler reproduces the training points.
This agrees with the memorisation of the training data observed for diffusion models at the empirical optimum~\citep{gu2023memorization, biroli2024dynamical}.

\section{Conclusion}
While deterministic samplers in diffusion models have largely fallen outside the variational framework, we have defined a deterministic model, the Degenerate Diffusion Model, and shown that it can be trained with a suitable variational method.
Applying the spread divergence at each step, we derived a well-defined objective that equals, up to a parameter-independent constant, the spread divergence between the generator and the backward distribution.
Its minimiser, the mean of the backward distribution, does not depend on the spread variance, so any fixed positive weighting of the per-step loss terms is itself a spread divergence.
We proved, in discrete time and without continuous-time arguments, that for a finite training set, the deterministic sampler at the optimum converges weakly to the data distribution as the number of steps tends to infinity.
Finally, although derived from the variational bound alone, the Degenerate Diffusion Model is equivalent to a discrete-time implementation of flow matching for a specific choice of spread, exactly so for the linear schedule, which gives the flow matching loss and its weightings a variational interpretation.

\section*{Acknowledgments}
This work was supported by the Engineering and Physical Sciences Research Council and AI Hub in Generative Models [grant numbers EP/S021566/1, EP/Y028805/1, UKRI5028].

\section*{Use of Large Language Models}
First, for Theorem~\ref{thm:marginals}, the authors formulated the result and devised the proof strategy and its key arguments, and Claude helped turn this informal argument into the detailed formal proof given in Appendix~\ref{app:convergence}.
All research ideas, results and claims are the authors' own. The authors checked every step of the proof and take full responsibility for the content of the paper.

\bibliography{tmlr}

@inproceedings{eijkelboom2024vfm,
  title     = {{Variational Flow Matching for Graph Generation}},
  author    = {Eijkelboom, Floor and Bartosh, Grigory and Naesseth, Christian A. and Welling, Max and van de Meent, Jan-Willem},
  booktitle = {Advances in Neural Information Processing Systems},
  volume    = {37},
  year      = {2024}
}

@book{rudin1987real,
  title     = {{Real and Complex Analysis}},
  author    = {Rudin, Walter},
  edition   = {3},
  year      = {1987},
  publisher = {McGraw-Hill}
}

@inproceedings{
lipman2023flowmatchinggenerativemodeling,
title={{Flow Matching for Generative Modeling}},
author={Yaron Lipman and Ricky T. Q. Chen and Heli Ben-Hamu and Maximilian Nickel and Matthew Le},
booktitle={The Eleventh International Conference on Learning Representations },
year={2023},
url={https://openreview.net/forum?id=PqvMRDCJT9t}
}

@inproceedings{
song2021ddim,
title={{Denoising Diffusion Implicit Models}},
author={Jiaming Song and Chenlin Meng and Stefano Ermon},
booktitle={International Conference on Learning Representations},
year={2021},
url={https://openreview.net/forum?id=St1giarCHLP}
}

@inproceedings{
song2021scoresde,
title={{Score-Based Generative Modeling through Stochastic Differential Equations}},
author={Yang Song and Jascha Sohl-Dickstein and Diederik P Kingma and Abhishek Kumar and Stefano Ermon and Ben Poole},
booktitle={International Conference on Learning Representations},
year={2021},
url={https://openreview.net/forum?id=PxTIG12RRHS}
}

@inproceedings{
liu2023rectifiedflow,
title={{Flow Straight and Fast: Learning to Generate and Transfer Data with Rectified Flow}},
author={Xingchao Liu and Chengyue Gong and qiang liu},
booktitle={The Eleventh International Conference on Learning Representations },
year={2023},
url={https://openreview.net/forum?id=XVjTT1nw5z}
}

@InProceedings{spreaddivergence,
  title = 	 {{Spread Divergence}},
  author =       {Zhang, Mingtian and Hayes, Peter and Bird, Thomas and Habib, Raza and Barber, David},
  booktitle = 	 {Proceedings of the 37th International Conference on Machine Learning},
  pages = 	 {11106--11116},
  year = 	 {2020},
  editor = 	 {III, Hal Daumé and Singh, Aarti},
  volume = 	 {119},
  series = 	 {Proceedings of Machine Learning Research},
  month = 	 {13--18 Jul},
  publisher =    {PMLR},
  url = 	 {https://proceedings.mlr.press/v119/zhang20j.html},
}

@inproceedings{
kingma2023understanding,
title={{Understanding Diffusion Objectives as the {ELBO} with Simple Data Augmentation}},
author={Diederik P Kingma and Ruiqi Gao},
booktitle={Thirty-seventh Conference on Neural Information Processing Systems},
year={2023},
url={https://openreview.net/forum?id=NnMEadcdyD}
}

@article{
benton2024error,
title={{Error Bounds for Flow Matching Methods}},
author={Joe Benton and George Deligiannidis and Arnaud Doucet},
journal={Transactions on Machine Learning Research},
issn={2835-8856},
year={2024},
url={https://openreview.net/forum?id=uqQPyWFDhY},
note={}
}

@InProceedings{sohl2015deep,
  title = 	 {{Deep Unsupervised Learning using Nonequilibrium Thermodynamics}},
  author = 	 {Sohl-Dickstein, Jascha and Weiss, Eric and Maheswaranathan, Niru and Ganguli, Surya},
  booktitle = 	 {Proceedings of the 32nd International Conference on Machine Learning},
  pages = 	 {2256--2265},
  year = 	 {2015},
  editor = 	 {Bach, Francis and Blei, David},
  volume = 	 {37},
  series = 	 {Proceedings of Machine Learning Research},
  address = 	 {Lille, France},
  month = 	 {07--09 Jul},
  publisher =    {PMLR},
  url = 	 {https://proceedings.mlr.press/v37/sohl-dickstein15.html},
}

@inproceedings{ho2020ddpm,
 author = {Ho, Jonathan and Jain, Ajay and Abbeel, Pieter},
 booktitle = {Advances in Neural Information Processing Systems},
 editor = {H. Larochelle and M. Ranzato and R. Hadsell and M.F. Balcan and H. Lin},
 pages = {6840--6851},
 publisher = {Curran Associates, Inc.},
 title = {{Denoising Diffusion Probabilistic Models}},
 url = {https://proceedings.neurips.cc/paper_files/paper/2020/file/4c5bcfec8584af0d967f1ab10179ca4b-Paper.pdf},
 volume = {33},
 year = {2020}
}

@misc{song2019generative,
      title={{Generative Modeling by Estimating Gradients of the Data Distribution}},
      author={Yang Song and Stefano Ermon},
      year={2020},
      eprint={1907.05600},
      archivePrefix={arXiv},
      primaryClass={cs.LG},
      url={https://arxiv.org/abs/1907.05600},
}

@inproceedings{dhariwal2021diffusion,
title={{Diffusion Models Beat {GAN}s on Image Synthesis}},
author={Prafulla Dhariwal and Alexander Quinn Nichol},
booktitle={Advances in Neural Information Processing Systems},
editor={A. Beygelzimer and Y. Dauphin and P. Liang and J. Wortman Vaughan},
year={2021},
url={https://openreview.net/forum?id=AAWuCvzaVt}
}

@misc{rombach2022ldm,
      title={{High-Resolution Image Synthesis with Latent Diffusion Models}},
      author={Robin Rombach and Andreas Blattmann and Dominik Lorenz and Patrick Esser and Björn Ommer},
      year={2022},
      eprint={2112.10752},
      archivePrefix={arXiv},
      primaryClass={cs.CV},
      url={https://arxiv.org/abs/2112.10752},
}

@inproceedings{
ho2022video,
title={{Video Diffusion Models}},
author={Jonathan Ho and Tim Salimans and Alexey A. Gritsenko and William Chan and Mohammad Norouzi and David J. Fleet},
booktitle={ICLR Workshop on Deep Generative Models for Highly Structured Data},
year={2022},
url={https://openreview.net/forum?id=BBelR2NdDZ5}
}

@article{blattmann2023svd,
  publtype={informal},
  author={Andreas Blattmann and Tim Dockhorn and Sumith Kulal and Daniel Mendelevitch and Maciej Kilian and Dominik Lorenz and Yam Levi and Zion English and Vikram Voleti and Adam Letts and Varun Jampani and Robin Rombach},
  title={{Stable Video Diffusion: Scaling Latent Video Diffusion Models to Large Datasets}},
  year={2023},
  cdate={1672531200000},
  journal={CoRR},
  volume={abs/2311.15127},
  url={https://doi.org/10.48550/arXiv.2311.15127}
}

@inproceedings{
kingma2021vdm,
title={{Variational Diffusion Models}},
author={Diederik P Kingma and Tim Salimans and Ben Poole and Jonathan Ho},
booktitle={Advances in Neural Information Processing Systems},
editor={A. Beygelzimer and Y. Dauphin and P. Liang and J. Wortman Vaughan},
year={2021},
url={https://openreview.net/forum?id=2LdBqxc1Yv}
}

@inproceedings{
karras2022edm,
title={{Elucidating the Design Space of Diffusion-Based Generative Models}},
author={Tero Karras and Miika Aittala and Timo Aila and Samuli Laine},
booktitle={Advances in Neural Information Processing Systems},
editor={Alice H. Oh and Alekh Agarwal and Danielle Belgrave and Kyunghyun Cho},
year={2022},
url={https://openreview.net/forum?id=k7FuTOWMOc7}
}

@inproceedings{
albergo2023building,
title={{Building Normalizing Flows with Stochastic Interpolants}},
author={Michael Samuel Albergo and Eric Vanden-Eijnden},
booktitle={The Eleventh International Conference on Learning Representations },
year={2023},
url={https://openreview.net/forum?id=li7qeBbCR1t}
}

@inproceedings{
esser2024sd3,
title={{Scaling Rectified Flow Transformers for High-Resolution Image Synthesis}},
author={Patrick Esser and Sumith Kulal and Andreas Blattmann and Rahim Entezari and Jonas M{\"u}ller and Harry Saini and Yam Levi and Dominik Lorenz and Axel Sauer and Frederic Boesel and Dustin Podell and Tim Dockhorn and Zion English and Robin Rombach},
booktitle={Forty-first International Conference on Machine Learning},
year={2024},
url={https://openreview.net/forum?id=FPnUhsQJ5B}
}

@inproceedings{
salimans2022progressive,
title={{Progressive Distillation for Fast Sampling of Diffusion Models}},
author={Tim Salimans and Jonathan Ho},
booktitle={International Conference on Learning Representations},
year={2022},
url={https://openreview.net/forum?id=TIdIXIpzhoI}
}

@inproceedings{
gao2025diffusionflow,
title={{Diffusion Models and Gaussian Flow Matching: Two Sides of the Same Coin}},
author={Ruiqi Gao and Emiel Hoogeboom and Jonathan Heek and Valentin De Bortoli and Kevin Patrick Murphy and Tim Salimans},
booktitle={The Fourth Blogpost Track at ICLR 2025},
year={2025},
url={https://openreview.net/forum?id=C8Yyg9wy0s}
}

@inproceedings{
chen2023probabilityflow,
title={{The Probability Flow {ODE} is Provably Fast}},
author={Sitan Chen and Sinho Chewi and Holden Lee and Yuanzhi Li and Jianfeng Lu and Adil Salim},
booktitle={Thirty-seventh Conference on Neural Information Processing Systems},
year={2023},
url={https://openreview.net/forum?id=KD6MFeWSAd}
}

@inproceedings{
li2023faster,
title={{Towards Non-Asymptotic Convergence for Diffusion-Based Generative Models}},
author={Gen Li and Yuting Wei and Yuxin Chen and Yuejie Chi},
booktitle={The Twelfth International Conference on Learning Representations},
year={2024},
url={https://openreview.net/forum?id=4VGEeER6W9}
}

@inproceedings{
gao2024wasserstein,
title={{Convergence Analysis for General Probability Flow {ODE}s of Diffusion Models in Wasserstein Distances}},
author={Xuefeng Gao and Lingjiong Zhu},
booktitle={The 28th International Conference on Artificial Intelligence and Statistics},
year={2025},
url={https://openreview.net/forum?id=EkO8rb3liX}
}

@article{
gu2023memorization,
title={{On Memorization in Diffusion Models}},
author={Xiangming Gu and Chao Du and Tianyu Pang and Chongxuan Li and Min Lin and Ye Wang},
journal={Transactions on Machine Learning Research},
issn={2835-8856},
year={2025},
url={https://openreview.net/forum?id=D3DBqvSDbj},
note={}
}

@article{biroli2024dynamical,
   title={{Dynamical Regimes of Diffusion Models}},
   volume={15},
   ISSN={2041-1723},
   url={http://dx.doi.org/10.1038/s41467-024-54281-3},
   DOI={10.1038/s41467-024-54281-3},
   number={1},
   journal={Nature Communications},
   publisher={Springer Science and Business Media LLC},
   author={Biroli, Giulio and Bonnaire, Tony and de Bortoli, Valentin and Mézard, Marc},
   year={2024},
   month=Nov }
\bibliographystyle{tmlr}

\appendix

\section{Standard Diffusion Bound}
\label{app:diffu}
\begin{lemma}[Standard diffusion bound]
  \label{th:diffu-loss}
  For the noise model $q(x_{1:T} \mid x_0) = \prod_{t=1}^T q(x_t \mid x_{t-1},x_0)$ and the Markovian
  generator $p_\theta(x_{0:T}) = p(x_T)\prod_{t=1}^T p_\theta(x_{t-1} \mid x_t)$ of Section~\ref{sec:bg-diffusion}, the bound
  \eqref{eq:diffu-loss} holds, with
  $C = \av{\mathrm{KL}[q(x_T \mid x_0) \,\|\, p(x_T)]}_{q(x_0)}$ independent of $\theta$.
\end{lemma}

\begin{proof}
  Fix $x_0$. Since $p_\theta(x_0) = \int dx_{1:T}\, p_\theta(x_{0:T})$, Jensen's inequality applied to the concave logarithm gives
  \begin{align}
    \log p_\theta(x_0)
    = \log \av{\frac{p_\theta(x_{0:T})}{q(x_{1:T} \mid x_0)}}_{q(x_{1:T} \mid x_0)}
    \ge \av{\log \frac{p_\theta(x_{0:T})}{q(x_{1:T} \mid x_0)}}_{q(x_{1:T} \mid x_0)}.
    \label{eq:proof-elbo}
  \end{align}
  Because the noise model is Markov given $x_0$, Bayes' rule gives the backward factorisation
  $q(x_{1:T} \mid x_0) = q(x_T \mid x_0) \prod_{t=2}^T q(x_{t-1} \mid x_t, x_0)$.
  Substituting it and $p_\theta(x_{0:T}) = p(x_T)\, p_\theta(x_0 \mid x_1) \prod_{t=2}^T p_\theta(x_{t-1} \mid x_t)$ into \eqref{eq:proof-elbo}, and grouping the terms that share the same variables, gives
  \begin{multline}
    -\log  p_\theta(x_0)
      \le \sum_{t=2}^T
        \av{\mathrm{KL}[q( x_{t-1} \mid x_t, x_0) \,\|\, p_\theta(x_{t-1} \mid x_t) ]}_{q(x_t \mid x_0)} \\
      + \mathrm{KL}[q(x_T \mid x_0) \,\|\, p(x_T)]
      - \av{\log p_\theta(x_0 \mid x_1)}_{q(x_1 \mid x_0)}.
  \label{eq:proof-decomposed}
  \end{multline}
  Averaging both sides over $q(x_0)$ gives \eqref{eq:diffu-loss}, with $C$ as stated. The term $C$ does not depend on $\theta$ because neither $q$ nor $p(x_T)$ does.
\end{proof}

\section{Proof of Lemma~\ref{th:xtgivenx} (Forward Probabilities)}
\label{lemma1_proofs}
\xtgivenx*
\begin{proof}
  \textbf{Marginal.} Conditioned on $x_0$, $x_t$ is an affine function of the Gaussian
  variable $\epsilon$, so $q(x_t \mid x_0)$ is Gaussian and is fully specified by its first
  two moments. Using $\av{\epsilon}=0$ and $\av{\epsilon\epsilon^\top}=I$,
  \begin{align}
    \fmu_t &= \av{x_t} = (1-\lambda_t)x_0 + \lambda_t\av{\epsilon} = (1-\lambda_t)x_0, \\
    \fSigma_t &= \av{(x_t-\fmu_t)(x_t-\fmu_t)^\top}
      = \lambda_t^2\av{\epsilon\epsilon^\top} = \lambda_t^2I.
  \end{align}

  \textbf{Posterior.} Since $x_{t-1}$ and $x_t$ are driven by the same $\epsilon$ and
  $\lambda_t > \lambda_0 = 0$ for $t \ge 1$, the forward equation can be inverted for the noise:
  \begin{align}
    \epsilon = \frac{x_t - (1-\lambda_t)x_0}{\lambda_t}.
  \end{align}
  
  Substituting into $x_{t-1} = (1-\lambda_{t-1})x_0 + \lambda_{t-1}\epsilon$ gives
  \begin{align}
    x_{t-1}
    &= (1-\lambda_{t-1})x_0 + \frac{\lambda_{t-1}}{\lambda_t}\br{x_t - (1-\lambda_t)x_0} \\
    &= \sq{1-\lambda_{t-1} - \frac{\lambda_{t-1}}{\lambda_t} + \lambda_{t-1}}x_0 + \frac{\lambda_{t-1}}{\lambda_t}x_t \\
    &= \br{1-\frac{\lambda_{t-1}}{\lambda_t}}x_0 + \frac{\lambda_{t-1}}{\lambda_t}x_t = \bmu_{t-1}(x_t, x_0).
  \end{align}
  Hence, given $(x_t,x_0)$, $x_{t-1}$ is determined exactly, i.e.,
  $q(x_{t-1} \mid x_t,x_0) = \delta\big(x_{t-1} \bigm|\bmu_{t-1}(x_t, x_0)\big)$.
\end{proof}

\section{Proof of Lemma~\ref{th:spread-decomp} (Decomposition of the Spread Objective)}
\label{app:proofs}

\spreaddecomp*

\begin{proof}
  Each term of $L'$ is an average over $q(x_0)q(x_t \mid x_0) = q(x_t)q(x_0 \mid x_t)$.
  We write $\tilde{q}(y_{t-1}, x_0 \mid x_t) := q(x_0 \mid x_t)\, \tilde{q}(y_{t-1} \mid x_t, x_0)$ for the joint of $x_0$ and the spread variable given $x_t$, whose marginal in $y_{t-1}$ is $\tilde{q}(y_{t-1} \mid x_t)$.
  Adding and subtracting $\log \tilde{q}(y_{t-1} \mid x_t)$ inside the divergence splits it in two:
  \begin{align}
    & \av{ \mathrm{KL}[\tilde{q}(y_{t-1} \mid x_t,x_0) \,\|\, \tilde{p}_\theta(y_{t-1} \mid x_t)] }_{q(x_t \mid x_0)q(x_0)} \nonumber \\
    &\quad = \av{ \int dy_{t-1} \, \tilde{q}(y_{t-1} \mid x_t,x_0)
        \log \frac{\tilde{q}(y_{t-1} \mid x_t)}{\tilde{p}_\theta(y_{t-1} \mid x_t)} }_{q(x_0 \mid x_t)q(x_t)} \nonumber \\
    &\hspace{5cm} + \av{ \int dy_{t-1} \, \tilde{q}(y_{t-1} \mid x_t,x_0)
        \log \frac{\tilde{q}(y_{t-1} \mid x_t,x_0)}{\tilde{q}(y_{t-1} \mid x_t)} }_{q(x_0 \mid x_t)q(x_t)} \label{eq:26} \\
    &\quad =  \av{ \int dy_{t-1} dx_0 \, \tilde{q}(y_{t-1}, x_0 \mid x_t)
        \log \frac{\tilde{q}(y_{t-1} \mid x_t)}{\tilde{p}_\theta(y_{t-1} \mid x_t)} }_{q(x_t)} + G_t \\
    &\quad = \av{\mathrm{KL}[\tilde{q}(y_{t-1} \mid x_t) \,\|\, \tilde{p}_\theta(y_{t-1} \mid x_t)]}_{q(x_t)} + G_t.
     \label{eq:28}
  \end{align}
  Here $G_t$ denotes the second term of \eqref{eq:26},
  \begin{align}
    G_t = \av{\mathrm{KL}[\tilde{q}(y_{t-1} \mid x_t,x_0) \,\|\, \tilde{q}(y_{t-1} \mid x_t)]}_{q(x_t \mid x_0) q(x_0)}.
  \end{align}
  The resulting identity between the first and the last line is the chain rule of the KL divergence for the joint $\tilde{q}(y_{t-1}, x_0 \mid x_t)$. 


Summing over $t$ gives $L'(\theta) = \mathcal{D}(\theta) + C$ with $C = \sum_{t=1}^T G_t$, which does not involve $\theta$.
\end{proof}

\section{Proof of Theorem~\ref{thm:marginals} (Convergence of the Deterministic Sampler)}
\label{app:convergence}

\textbf{Notation.} As in Theorem~\ref{thm:marginals}, $\hat{x}_T, \hat{x}_{T-1}, \ldots, \hat{x}_0$ are the states of the sampler, and $\hat{x}_0$ is the generated sample. Without a hat, $x_t$ is the forward variable, with marginal $q_t$, and $x_0^1, \ldots, x_0^N$ are the training points. The sampler uses the forward probabilities only through $q_t$ and the related functions $m_t$ and $w_n^t$ of Section~\ref{sec:convergence} (recalled below), which we evaluate at generic points $x \in \R^d$. We also use a reference chain $y_T, \ldots, y_0$. This $y$ is unrelated to the spread variable $y_{t-1}$ of Section~\ref{sec:spread-obj} and Appendix~\ref{app:proofs}.

\textbf{Overview.} The proof has four parts.
(i) We define the reference chain, which starts from the same noise as the sampled chain but samples from the backward distribution, and we choose the constants $R$, $b$, $\rho$, $\eta$ and $\zeta$.
(ii) Lemma~\ref{lem:phase1} (Tracking) handles the high-noise steps, $\lambda_t \ge \zeta$. It shows that the reference chain has the forward marginals, $y_t \sim q_t$, and that at the end of this regime the expected squared distance between the sampled chain $\hat{x}_t$ and the reference chain $y_t$ is $O(h)$ for fixed $\zeta$, where $h$ is the largest step size.
(iii) Lemma~\ref{lem:trapping} (Trapping) handles the low-noise steps, $\lambda_t \le 2\zeta$. Each training point $x_0^m$ gives one Gaussian component of $q_t$, with centre $(1-\lambda_t)x_0^m$. Once the sampled chain is within a small distance ($\rho$) of this centre, it stays within $\rho$ of it at every later step, and the generated sample $\hat{x}_0$ ends within $\gamma_T = R\,\eta(\lambda_1)$ of $x_0^m$.
(iv) The proof of Theorem~\ref{thm:marginals} joins the two regimes at $\tau$, the smallest $t$ with $\lambda_t \ge \zeta$, i.e., the last high-noise step. The marginal $q_\tau$ is a mixture of $N$ Gaussians with weights $\frac{1}{N}$, one per training point $x_0^n$, centred at $(1-\lambda_\tau)x_0^n$ with standard deviation $\lambda_\tau$, and $\lambda_\tau < 2\zeta$ once $T$ is large enough such that $h \le \zeta$. Hence the reference chain $y_\tau \sim q_\tau$ lies within $\rho/2$ of the centre $(1-\lambda_\tau)x_0^m$ of training point $x_0^m$ with probability close to $\frac{1}{N}$ when $\zeta$ is small. By (ii), the sampled chain $\hat{x}_\tau$ is within $\rho/2$ of $y_\tau$ with probability close to one, so it is within $\rho$ of the centre and (iii) applies. This gives a lower bound on $P(\norm{\hat{x}_0 - x_0^m} \le \gamma_T)$, and, since the balls around the training points are disjoint, applying it to the other $N-1$ points gives an upper bound. Both tend to $\frac{1}{N}$ as $T \to \infty$ and $\zeta \to 0$. Weak convergence follows.

\textbf{Definitions.} The sampled chain $\hat{x}_T, \hat{x}_{T-1}, \ldots, \hat{x}_0$ starts at $\hat{x}_T \sim \ndist{\cdot}{0}{I}$, and its marginals are unknown. We compare it with a reference chain $y_T, y_{T-1}, \ldots, y_0$, defined in \eqref{eq:reference-chain} below, whose marginals are the known $q_t$ (Lemma~\ref{lem:phase1}).
We define the maximum and minimum distance between two training points as
\begin{align}
R = \max_{n,m}\norm{x_0^n - x_0^m}, \qquad b = \min_{n \neq m} \norm{x_0^n - x_0^m} >0,
\label{eq:Rb}
\end{align}
where $b$ is well-defined and positive because $N \ge 2$ and the points are distinct.
From them we define a radius $\rho$ and the function $\eta$ on $\lambda > 0$, which is increasing in $\lambda$ and vanishes as $\lambda \to 0$,
\begin{align}
\rho = \frac{b}{8}, \qquad \eta(\lambda) = (N-1)\exp\left\{-\frac{b^2}{16\lambda^2}\right\}.
\end{align}
Throughout this appendix, including Lemmas~\ref{lem:phase1} and~\ref{lem:trapping}, we assume as in Theorem~\ref{thm:marginals} that the generator is at the exact optimum of the spread objective, $\mu_\theta(x) = \mu_t^*(x)$ for every $x \in \R^d$. Since $\bmu_{t-1}(x, x_0) = x - \frac{\Delta\lambda_t}{\lambda_t}(x - x_0)$, this optimum can be written as $\mu_t^*(x) = x - \Delta\lambda_t\, v_t^*(x)$ with $v_t^*(x) = (x - m_t(x))/\lambda_t$, and we work with this velocity form throughout. We recall from Section~\ref{sec:convergence} the posterior mean $m_t$ of the data under $q_t$ and the posterior weights $w_n^t$, now with their normaliser written out; $K$ denotes the index of the training point that generated $x_t$:
\begin{align}
m_t(x) &= \av{x_0}_{q(x_0 \mid x_t = x)} = \sum_{n=1}^N w_n^t(x)\, x_0^n, \\
w_n^t(x) &= q(K = n \mid x_t = x) = \frac{\exp\left\{-\norm{x - (1-\lambda_t)x_0^n}^2 / (2\lambda_t^2)\right\}}{\sum_{k=1}^N \exp\left\{-\norm{x - (1-\lambda_t)x_0^k}^2 / (2\lambda_t^2)\right\}}.
\end{align}
With $r_t = \lambda_{t-1}/\lambda_t$, so that $\bmu_{t-1}(x, x_0) = r_t\, x + (1-r_t)\, x_0$ by Lemma~\ref{th:xtgivenx}, the sampler update $\hat{x}_{t-1} = \hat{x}_t - \Delta\lambda_t\, v_\theta(\hat{x}_t)$ then reads
\begin{align}
\hat{x}_{t-1} = \bmu_{t-1}(\hat{x}_t, m_t(\hat{x}_t)) = r_t\, \hat{x}_t + (1-r_t)\, m_t(\hat{x}_t). \label{eq:sampler-mean-form}
\end{align}
We define the reference chain to start from the same noise, $y_T = \hat{x}_T$, and sample at each step one training point from the posterior instead of averaging over it,
\begin{align}
y_{t-1} = \bmu_{t-1}(y_t, x_0^{J_t}) = r_t\, y_t + (1-r_t)\, x_0^{J_t}, \qquad P(J_t = n \mid \gF_t) = w_n^t(y_t), \label{eq:reference-chain}
\end{align}
where $\gF_t$ is the $\sigma$-algebra generated by $(\hat{x}_T, J_T, \ldots, J_{t+1})$, the past of the reference chain, which determines $y_t$. It also determines the state $\hat{x}_t$ of the sampled chain, which is a deterministic function of $\hat{x}_T$.
With the step sizes $\Delta\lambda_t$ of Section~\ref{sec:reparam}, $1 - r_t = \Delta\lambda_t/\lambda_t$, and we write $h = \max_t \Delta\lambda_t$ for the largest step.
Finally, we fix a constant $\zeta > 0$ small enough that
\begin{align}
\zeta \le \frac{1}{4} \quad \text{and} \quad R\,\eta(2\zeta) \le \rho. \label{eq:delta-choice}
\end{align}
Such a $\zeta$ exists because $\eta(\lambda) \to 0$ as $\lambda \to 0$, and, since $\eta$ is increasing, every smaller $\zeta$ satisfies \eqref{eq:delta-choice} as well.

\begin{restatable}[Tracking]{lemma}{tracking}\label{lem:phase1}
Let $\tau = \min\{t : \lambda_t \ge \zeta\}$. The reference chain \eqref{eq:reference-chain} has the forward marginals, $y_t \sim q_t$ for every $t$, and the sampled chain stays close to it at time $\tau$:
\begin{align}
\av{\norm{\hat{x}_\tau - y_\tau}^2} \le \exp\br{\frac{2R^2}{\zeta^3}}\, \frac{R^2 h}{\zeta^2}.
\end{align}
Moreover, $\lambda_\tau < 2\zeta$ whenever $h \le \zeta$.
\end{restatable}
\begin{proof}
\textbf{Reference chain marginals.} By Lemma~\ref{th:xtgivenx}, $q(x_{t-1} \mid x_t, x_0) = \delta\big(x_{t-1} \mid \bmu_{t-1}(x_t, x_0)\big)$, so the backward distribution is
\begin{align}
q(x_{t-1} \mid x_t) = \sum_{n=1}^N q(K = n \mid x_t)\, q(x_{t-1} \mid x_t, x_0^n)
= \sum_{n=1}^N w_n^t(x_t)\, \delta\big(x_{t-1} \mid \bmu_{t-1}(x_t, x_0^n)\big).
\end{align}
A step of the reference chain \eqref{eq:reference-chain} is exactly a draw $y_{t-1} \sim q(x_{t-1} \mid x_t = y_t)$. Hence, if $y_t \sim q_t$, then $y_{t-1}$ has density
\begin{align}
\int q(x_{t-1} \mid x_t)\, q_t(x_t)\, dx_t = q_{t-1}(x_{t-1}),
\end{align}
and since $y_T \sim \ndist{\cdot}{0}{I} = q_T$, induction downwards gives $y_t \sim q_t$ for every $t$.

\textbf{Lipschitz bound of posterior mean.} Let $C_t(x) = \sum_n w_n^t(x)\,(x_0^n - m_t(x))(x_0^n - m_t(x))^\top$ be the posterior covariance of the data given $x_t = x$. Differentiating the weights gives the gradient $\nabla w_n^t(x) = \frac{1-\lambda_t}{\lambda_t^2}\, w_n^t(x)\,(x_0^n - m_t(x))$, hence the Jacobian of $m_t$ is
\begin{align}
\nabla m_t(x) = \sum_{n=1}^N x_0^n\, \nabla w_n^t(x)^\top = \frac{1-\lambda_t}{\lambda_t^2}\, C_t(x).
\end{align}
As $m_t(x)$ is a convex combination of the training points, $\norm{x_0^n - m_t(x)} \le R$. Hence, for any vector $\nu \in \R^d$, by the triangle and Cauchy--Schwarz inequalities,
\begin{align}
\norm{C_t(x)\, \nu}
= \norm{\sum_{n=1}^N w_n^t(x)\, (x_0^n - m_t(x))\, (x_0^n - m_t(x))^\top \nu}
\le \sum_{n=1}^N w_n^t(x) \norm{x_0^n - m_t(x)}^2 \norm{\nu} \le R^2 \norm{\nu}. \label{eq:trace-bound}
\end{align}
Now fix $x, x' \in \R^d$ and let $x_\alpha = x' + \alpha(x - x')$, $\alpha \in [0,1]$, be the segment from $x'$ to $x$. By the fundamental theorem of calculus and the chain rule,
\begin{align}
m_t(x) - m_t(x') = \int_0^1 \frac{d}{d\alpha}\, m_t(x_\alpha)\, d\alpha
= \int_0^1 \nabla m_t(x_\alpha)\, (x - x')\, d\alpha
= \frac{1-\lambda_t}{\lambda_t^2} \int_0^1 C_t(x_\alpha)\, (x - x')\, d\alpha.
\end{align}
Taking norms, applying \eqref{eq:trace-bound} with $\nu = x - x'$ at every point $x_\alpha$, and using $1 - \lambda_t \le 1$,
\begin{align}
\norm{m_t(x) - m_t(x')}
\le \frac{1-\lambda_t}{\lambda_t^2} \int_0^1 \norm{C_t(x_\alpha)\, (x - x')}\, d\alpha
\le \frac{1-\lambda_t}{\lambda_t^2} \int_0^1 R^2 \norm{x - x'}\, d\alpha
\le \frac{R^2}{\lambda_t^2} \norm{x - x'}. \label{eq:lipschitz}
\end{align}

\textbf{Error recursion.} Let $e_t = \hat{x}_t - y_t$, so $e_T = 0$, and let $\xi_t = m_t(y_t) - x_0^{J_t}$ be the difference between averaging and picking. Subtracting \eqref{eq:reference-chain} from \eqref{eq:sampler-mean-form}, and adding and subtracting $(1-r_t)\, m_t(y_t)$,
\begin{align}
e_{t-1} = a_t + (1-r_t)\, \xi_t, \qquad a_t = r_t\, e_t + (1-r_t)\big(m_t(\hat{x}_t) - m_t(y_t)\big). \label{eq:error-recursion}
\end{align}

The term $a_t$ is fixed given $\gF_t$, since $\hat{x}_t$ and $y_t$ are, and the only remaining randomness at step $t$ is $J_t$, which enters only through $\xi_t$. Given $\gF_t$, $J_t$ is drawn from the weights $(w_n^t(y_t))_{n=1}^N$, so, using $\norm{x_0^n - m_t(y_t)} \le R$ and $\sum_{n=1}^N w_n^t(y_t) = 1$,
\begin{align}
\av{\xi_t}_{P(\cdot \mid \gF_t)} &= m_t(y_t) - \sum_{n=1}^N w_n^t(y_t)\, x_0^n = 0, \nonumber \\
\av{\norm{\xi_t}^2}_{P(\cdot \mid \gF_t)} &= \sum_{n=1}^N w_n^t(y_t) \norm{m_t(y_t) - x_0^n}^2 \le R^2. \label{eq:kick}
\end{align}

\textbf{One step.} Let $V_t = \av{\norm{e_t}^2}$. Squaring \eqref{eq:error-recursion}, the cross term vanishes by \eqref{eq:kick}, since $\av{a_t^\top \xi_t} = \av{a_t^\top\, \av{\xi_t}_{P(\cdot \mid \gF_t)}} = 0$. Using $1 - r_t = \Delta\lambda_t/\lambda_t$,
\begin{align}
V_{t-1} = \av{\norm{a_t}^2} + \frac{(\Delta\lambda_t)^2}{\lambda_t^2}\, \av{\norm{\xi_t}^2} \le \av{\norm{a_t}^2} + \frac{R^2 (\Delta\lambda_t)^2}{\lambda_t^2}.
\end{align}
By \eqref{eq:lipschitz} and $r_t \le 1$,
\begin{align}
\norm{a_t} \le r_t \norm{e_t} + \frac{\Delta\lambda_t}{\lambda_t}\, \frac{R^2}{\lambda_t^2} \norm{e_t} \le (1 + \kappa_t) \norm{e_t}, \qquad \kappa_t = \frac{R^2 \Delta\lambda_t}{\lambda_t^3},
\end{align}
and therefore
\begin{align}
V_{t-1} \le (1+\kappa_t)^2\, V_t + \frac{R^2 (\Delta\lambda_t)^2}{\lambda_t^2}. \label{eq:V-recursion}
\end{align}

\textbf{Unrolling.} Applying \eqref{eq:V-recursion} from $V_T = 0$ down to $\tau$ and using $1 + \kappa \le e^\kappa$ for $\kappa \ge 0$,
\begin{align}
V_\tau \le \sum_{t=\tau+1}^T \br{\prod_{s=\tau+1}^{t-1} (1+\kappa_s)^2} \frac{R^2 (\Delta\lambda_t)^2}{\lambda_t^2}
\le \exp\br{2 \sum_{t=\tau+1}^T \kappa_t}\, \sum_{t=\tau+1}^T \frac{R^2 (\Delta\lambda_t)^2}{\lambda_t^2}.
\end{align}
Since $\lambda$ is increasing, $\lambda_t \ge \lambda_\tau \ge \zeta$ on these steps. With $\sum_t \Delta\lambda_t = \lambda_T - \lambda_0 = 1$ and $\Delta\lambda_t \le h$,
\begin{align}
\sum_{t=\tau+1}^T \kappa_t \le \frac{R^2}{\zeta^3} \sum_{t=\tau+1}^T \Delta\lambda_t \le \frac{R^2}{\zeta^3},
\qquad
\sum_{t=\tau+1}^T \frac{R^2 (\Delta\lambda_t)^2}{\lambda_t^2} \le \frac{R^2 h}{\zeta^2} \sum_{t=\tau+1}^T \Delta\lambda_t \le \frac{R^2 h}{\zeta^2},
\end{align}
which gives the bound on $V_\tau = \av{\norm{\hat{x}_\tau - y_\tau}^2}$.

\textbf{Time $\tau$.} Since $\lambda_0 = 0 < \zeta$, we have $\tau \ge 1$ and $\lambda_{\tau-1} < \zeta$ by the minimality of $\tau$. Hence $\lambda_\tau = \lambda_{\tau-1} + \Delta\lambda_\tau < \zeta + h \le 2\zeta$ when $h \le \zeta$.
\end{proof}

\begin{restatable}[Trapping]{lemma}{trapping}\label{lem:trapping}
If $\lambda_s \le 2\zeta$ and $\norm{\hat{x}_s - (1-\lambda_s)x_0^m} \le \rho$ for some $s \ge 1$ and some training point $x_0^m$, then the generated sample satisfies
\begin{align}
\norm{\hat{x}_0 - x_0^m} \le R\,\eta(\lambda_1).
\end{align}
\end{restatable}
\begin{proof}
\textbf{Peaking.} We first show that, for any $t$ with $\lambda_t \le 1/2$ and any $x$ with $\norm{x - (1-\lambda_t)x_0^m} \le \rho$,
\begin{align}
\norm{m_t(x) - x_0^m} \le R\,\eta(\lambda_t). \label{eq:peaking}
\end{align}
Fix $n \neq m$. By the definition of $b$ and $\lambda_t \le 1/2$, the two centres satisfy
\begin{align}
\norm{(1-\lambda_t)x_0^n - (1-\lambda_t)x_0^m} = (1-\lambda_t)\norm{x_0^n - x_0^m} \ge (1-\lambda_t)\, b \ge \frac{b}{2}.
\end{align}
By assumption, the distance of $x$ to its own centre is
\begin{align}
\norm{x - (1-\lambda_t)x_0^m} \le \rho = \frac{b}{8},
\end{align}
and hence, by the reverse triangle inequality, its distance to any other centre is
\begin{align}
\norm{x - (1-\lambda_t)x_0^n}
&\ge \norm{(1-\lambda_t)x_0^n - (1-\lambda_t)x_0^m} - \norm{x - (1-\lambda_t)x_0^m} \nonumber \\
&\ge \frac{b}{2} - \frac{b}{8} = \frac{3b}{8}.
\end{align}
The normaliser cancels in the ratio of the weights, so
\begin{align}
\frac{w_n^t(x)}{w_m^t(x)}
= \exp\left\{\frac{\norm{x - (1-\lambda_t)x_0^m}^2 - \norm{x - (1-\lambda_t)x_0^n}^2}{2\lambda_t^2}\right\}
\le \exp\left\{\frac{(b/8)^2 - (3b/8)^2}{2\lambda_t^2}\right\}
= \exp\left\{-\frac{b^2}{16\lambda_t^2}\right\}.
\end{align}
The weights sum to one, so $1 - w_m^t(x) = \sum_{n \neq m} w_n^t(x)$. Since $0 < w_m^t(x) \le 1$, dividing by it can only increase each term, $w_n^t(x) \le w_n^t(x)/w_m^t(x)$, and each of the $N-1$ ratios is bounded as above. Hence
\begin{align}
1 - w_m^t(x) = \sum_{n \neq m} w_n^t(x) \le \sum_{n \neq m} \frac{w_n^t(x)}{w_m^t(x)} \le (N-1)\exp\left\{-\frac{b^2}{16\lambda_t^2}\right\} = \eta(\lambda_t).
\end{align}
As the weights sum to one, $m_t(x) - x_0^m = \sum_{n \neq m} w_n^t(x)(x_0^n - x_0^m)$, which has norm at most $R(1 - w_m^t(x)) \le R\,\eta(\lambda_t)$. This proves \eqref{eq:peaking}.

\textbf{Trapping.} Let $\varphi_t =\norm{\hat{x}_t - (1-\lambda_t)x_0^m}$ be the distance of the sampled chain to the moving centre. Because $1 - \lambda_{t-1} = r_t(1-\lambda_t) + (1-r_t)$, subtracting $(1-\lambda_{t-1})x_0^m$ from both sides of \eqref{eq:sampler-mean-form} gives the exact identity
\begin{align}
\hat{x}_{t-1} - (1-\lambda_{t-1})x_0^m = r_t\big(\hat{x}_t - (1-\lambda_t)x_0^m\big) + (1-r_t)\big(m_t(\hat{x}_t) - x_0^m\big).
\end{align}
Take any $2 \le t \le s$ with $\varphi_t \le \rho$. Since $\lambda$ is increasing, $\lambda_t \le \lambda_s \le 2\zeta \le 1/2$ by $\zeta \le 1/4$, so \eqref{eq:peaking} applies at $x = \hat{x}_t$. As $r_t \in [0,1]$ and $\eta$ is increasing, the choice of $\zeta$ in \eqref{eq:delta-choice} gives
\begin{align}
\varphi_{t-1} \le r_t\, \varphi_t +(1-r_t)\, R\,\eta(\lambda_t) \le r_t\, \rho + (1-r_t)\, R\,\eta(2\zeta) \le \rho.
\end{align}
Since $\varphi_s \le \rho$ by assumption, induction gives $\varphi_t \le \rho$ for all $1 \le t \le s$. At the last step $r_1 = \lambda_0/\lambda_1 = 0$, so $\hat{x}_0 = m_1(\hat{x}_1)$, and \eqref{eq:peaking} with $\varphi_1 \le \rho$ gives $\norm{\hat{x}_0 - x_0^m} \le R\,\eta(\lambda_1)$.
\end{proof}
Each step moves the sampled chain part of the way towards a point within $\rho$ of the centre, so the trajectory never leaves the $\rho$-tube around $(1-\lambda_t)x_0^m$. The final step jumps onto the posterior mean, which lies within $R\,\eta(\lambda_1)$ of $x_0^m$, a distance exponentially small in $1/\lambda_1^2$.

\marginalsequal*
\begin{proof}
Our aim is to show that, starting at a random noise point, the sampled chain ends near each training example with probability tending to $1/N$. We do so in six steps: we set up two events at an intermediate time $\tau$, show that together they force the sampled chain to land next to $x_0^m$, bound the probability of landing there from below and from above, take the limit, and deduce weak convergence.

\textbf{Setup.} Let $\zeta$ be as in \eqref{eq:delta-choice} and $\tau = \min\{t : \lambda_t \ge \zeta\}$ as in Lemma~\ref{lem:phase1}. We take $T$ large enough that $h \le \zeta$, so that $\lambda_\tau < 2\zeta$ by that lemma. We set $\gamma_T = R\,\eta(\lambda_1)$ and define the ball around a data point $x_0^m$ as
\begin{align}
    B_m = \{x : \norm{x - x_0^m} \le \gamma_T\}.
\end{align}
With $\rho = b/8$, we consider two events at time $\tau$:
\begin{itemize}
\item $E_1$: the reference chain is near the centre of training point $m$, i.e., $\norm{y_\tau - (1-\lambda_\tau)x_0^m} \le \rho/2$;
\item $E_2$: the sampled chain is near the reference chain, i.e., $\norm{\hat{x}_\tau - y_\tau} \le \rho/2$.
\end{itemize}

\textbf{Inclusion.} We show that $E_1 \cap E_2 \subseteq \{\hat{x}_0 \in B_m\}$. If both events happen, the triangle inequality gives
\begin{align}
    \norm{\hat{x}_\tau - (1-\lambda_\tau)x_0^m} \le \norm{\hat{x}_\tau - y_\tau} + \norm{y_\tau - (1-\lambda_\tau)x_0^m} \le \frac{\rho}{2} + \frac{\rho}{2} = \rho.
\end{align}
Since also $\lambda_\tau < 2\zeta$, Lemma~\ref{lem:trapping} with $s = \tau$ gives $\norm{\hat{x}_0 - x_0^m} \le R\,\eta(\lambda_1) = \gamma_T$, that is $\hat{x}_0 \in B_m$.

\textbf{Lower bound.} By the inclusion, and since $P(E_1 \cap E_2) = P(E_1) - P(E_1 \cap E_2^c) \ge P(E_1) - P(E_2^c)$,
\begin{align}
P(\hat{x}_0 \in B_m) \ge P(E_1 \cap E_2) \ge P(E_1) - P(E_2^c). \label{eq:two-events}
\end{align}
We bound the two terms in turn, starting with $P(E_1)$.

Write $c = (1-\lambda_\tau)x_0^m$ and $S = \{y : \norm{y - c} \le \rho/2\}$,
so that $E_1 = \{y_\tau \in S\}$. By Lemma~\ref{lem:phase1}, $y_\tau$ has density
\[
  q_\tau(y) = \frac1N \sum_{n=1}^N \ndist{y}{(1-\lambda_\tau)x_0^n}{\lambda_\tau^2 I}
\]
Hence
\begin{align}
P(E_1) &= \int_S q_\tau(y)\,dy
        = \frac1N \sum_{n=1}^N \int_S \ndist{y}{(1-\lambda_\tau)x_0^n}{\lambda_\tau^2 I}\,dy \nonumber \\
       &= \frac1N \int_S \ndist{y}{c}{\lambda_\tau^2 I}\,dy
        + \frac1N \sum_{n \neq m} \underbrace{\int_S \ndist{y}{(1-\lambda_\tau)x_0^n}{\lambda_\tau^2 I}\,dy}_{\ge\, 0} \nonumber \\
       &\ge \frac1N \int_S \ndist{y}{c}{\lambda_\tau^2 I}\,dy.
\end{align}
The remaining integral is the probability that a variable with distribution $\ndist{\cdot}{c}{\lambda_\tau^2 I}$ lies in $S$. Such a variable can be written as $c + \lambda_\tau z$ with $z \sim \ndist{\cdot}{0}{I}$, and $c + \lambda_\tau z \in S$ exactly when $\norm{(c + \lambda_\tau z) - c} = \lambda_\tau \norm{z} \le \rho/2$. Hence
\begin{align}
\int_S \ndist{y}{c}{\lambda_\tau^2 I}\,dy
  = P\br{c + \lambda_\tau z \in S}
  = P\br{\lambda_\tau \norm{z} \le \rho/2}
  = 1 - P\br{\norm{z} > \frac{\rho}{2\lambda_\tau}}.
\end{align}
By Markov's inequality applied to $\norm{z}^2 \ge 0$,
\begin{align}
P\br{\norm{z} > \frac{\rho}{2\lambda_\tau}}
  = P\br{\norm{z}^2 > \frac{\rho^2}{4\lambda_\tau^2}}
  \le \frac{4\lambda_\tau^2\,\av{\norm{z}^2}}{\rho^2}
  = \frac{4\,\lambda_\tau^2\,d}{\rho^2}
  \le \frac{16\,d\,\zeta^2}{\rho^2},
\end{align}
using $\av{\norm{z}^2} = d$ and $\lambda_\tau < 2\zeta$. Therefore
\begin{align}
P(E_1) \;\ge\; \frac1N\br{1 - \frac{16\,d\,\zeta^2}{\rho^2}}.
\end{align}

For the second term, the complement of $E_2$ is $E_2^c = \{\norm{\hat{x}_\tau - y_\tau} > \rho/2\}
= \{\norm{\hat{x}_\tau - y_\tau}^2 > \rho^2/4\}$. By Markov's inequality applied to
$\norm{\hat{x}_\tau - y_\tau}^2 \ge 0$ and Lemma~\ref{lem:phase1} for the second inequality,
\begin{align}
  P(E_2^c) \;\le\; \frac{4\,\av{\norm{\hat{x}_\tau - y_\tau}^2}}{\rho^2}
          \;\le\; \frac{4}{\rho^2}\, \exp\br{\frac{2R^2}{\zeta^3}}\, \frac{R^2 h}{\zeta^2}.
\end{align}
Substituting the two bounds into \eqref{eq:two-events},
\begin{align}
  P(\hat{x}_0 \in B_m) \;\ge\; P(E_1) - P(E_2^c)
  &\ge \frac1N - \frac1N\,\frac{16\,d\,\zeta^2}{\rho^2}
     - \frac{4}{\rho^2}\, \exp\br{\frac{2R^2}{\zeta^3}}\, \frac{R^2 h}{\zeta^2} \nonumber \\
  &= \frac1N - \beta,
  \qquad
  \beta = \frac{16\,d\,\zeta^2}{N\rho^2}
     + \frac{4}{\rho^2}\, \exp\br{\frac{2R^2}{\zeta^3}}\, \frac{R^2 h}{\zeta^2}.
  \label{eq:lower-bound-beta}
\end{align}

\textbf{Upper bound.} The lower bound \eqref{eq:lower-bound-beta} holds for every data point $x_0^m$, and $\beta$ does not depend on the index $m$. Since $\gamma_T \to 0$ as $T \to \infty$, we have $\gamma_T < b/2$ for $T$ large enough, and then the balls are pairwise disjoint: a point in both $B_n$ and $B_m$ with $n \neq m$ would give $\norm{x_0^n - x_0^m} \le 2\gamma_T < b$, contradicting the definition of $b$ in \eqref{eq:Rb}. The events $\{\hat{x}_0 \in B_n\}$ are therefore mutually exclusive and their probabilities sum to at most one,
\begin{align}
  \sum_{n=1}^N P(\hat{x}_0 \in B_n) = P\br{\hat{x}_0 \in \bigcup_{n=1}^N B_n} \le 1.
\end{align}
Hence, applying the lower bound to each of the $N-1$ other points,
\begin{align}
  P(\hat{x}_0 \in B_m) \;\le\; 1 - \sum_{n \neq m} P(\hat{x}_0 \in B_n)
  \;\le\; 1 - (N-1)\br{\frac1N - \beta}
  \;=\; \frac1N + (N-1)\,\beta.
  \label{eq:upper-bound-beta}
\end{align}

\textbf{Limit.} Fix $\zeta$ and let $T \to \infty$, so that the largest step size $h \to 0$ by assumption. The second term of $\beta$ vanishes while the first does not depend on $T$, so
\begin{align}
  \lim_{T \to \infty} \beta = \frac{16\,d\,\zeta^2}{N\rho^2} =: \beta_\zeta.
\end{align}
Moreover, $\lambda_1 \le h \to 0$ gives $\gamma_T = R\,\eta(\lambda_1) \to 0$. Taking limits in \eqref{eq:lower-bound-beta} and \eqref{eq:upper-bound-beta},
\begin{align}
  \frac1N - \beta_\zeta
  \;\le\; \liminf_{T \to \infty} P(\hat{x}_0 \in B_m)
  \;\le\; \limsup_{T \to \infty} P(\hat{x}_0 \in B_m)
  \;\le\; \frac1N + (N-1)\,\beta_\zeta.
\end{align}
The event $\{\hat{x}_0 \in B_m\}$ does not depend on $\zeta$, so the two middle terms do not either, and the inequalities hold for every admissible $\zeta$. Letting $\zeta \to 0$ gives $\beta_\zeta \to 0$, and therefore
\begin{align}
  \lim_{T \to \infty} P(\hat{x}_0 \in B_m) = \frac1N.
\end{align}

\textbf{Weak convergence.} Let $f$ be a bounded continuous function and let $B = \bigcup_{n=1}^N B_n$. For $T$ large enough the balls are disjoint, so $P(\hat{x}_0 \in B) = \sum_{n=1}^N P(\hat{x}_0 \in B_n) \to 1$ and
\begin{align}
  \left| \av{f(\hat{x}_0)} - \sum_{n=1}^N f(x_0^n)\, P(\hat{x}_0 \in B_n) \right|
  \le \sum_{n=1}^N \sup_{x \in B_n} \left| f(x) - f(x_0^n) \right| P(\hat{x}_0 \in B_n)
  + \sup_{x} |f(x)|\; P(\hat{x}_0 \notin B).
\end{align}
The first term vanishes because $f$ is continuous at each $x_0^n$ and the radius $\gamma_T \to 0$, and the second because $P(\hat{x}_0 \notin B) \to 0$. Hence $\av{f(\hat{x}_0)} = \av{f(x_0)}_{p_\theta(x_0)} \to \frac1N \sum_{n=1}^N f(x_0^n) = \av{f(x_0)}_{q(x_0)}$, which is weak convergence of the distribution of $\hat{x}_0$ to $q(x_0)$.
\end{proof}

\section{Continuity of the Per-Time Loss}
\label{app:ell-continuous}
\begin{lemma}
  \label{th:ell-continuous}
  Let $\lambda : [0,1] \to [0,1]$ be continuous, let $v_\theta$ be continuous in $(x, s)$ with
  $\norm{v_\theta(x, s)} \le A + B\norm{x}^k$ for some integer $k \ge 1$, all $x$ and all $s \in [0,1]$, and let all moments of
  $q(x_0)$ be finite. Then $\ell$ in \eqref{eq:fm-loss2} is continuous on $[0,1]$, and hence bounded.
\end{lemma}
\begin{proof}
Since $x_s$ is a deterministic function of $(x_0, \epsilon)$, we can write
\begin{align}
  \ell(s) = \av{g(s; x_0, \epsilon)}_{q(x_0)\, q(\epsilon)},
  \qquad
  g(s; x_0, \epsilon) = \norm{\epsilon - x_0 - v_\theta(x_s, s)}^2,
\end{align}
with $x_s = (1-\lambda(s))x_0 + \lambda(s)\epsilon$ and $q(\epsilon) = \ndist{\epsilon}{0}{I}$.

\textbf{Continuity of the integrand ($g$).} Fix $(x_0, \epsilon)$. Since $\lambda$ is continuous, $x_s = (1-\lambda(s))x_0 + \lambda(s)\epsilon$ is continuous in $s$. Since $v_\theta$ is continuous in both of its arguments,\footnote{This is a mild assumption. Networks composed of standard layers (linear and convolutional layers, pointwise activations such as ReLU or SiLU, normalisation and attention) are continuous in their input. They are also continuous in $s$ when the time embedding is, as for the usual sinusoidal or learned embeddings of a real-valued $s$.} the network output $v_\theta(x_s, s)$ is then also continuous in $s$, and hence so is $g(s; x_0, \epsilon) = \norm{\epsilon - x_0 - v_\theta(x_s, s)}^2$.

\textbf{Domination.} By the triangle inequality, and since $1-\lambda(s)$ and $\lambda(s)$ both lie in $[0,1]$,
\begin{align}
  \norm{x_s} \le (1-\lambda(s))\norm{x_0} + \lambda(s)\norm{\epsilon} \le \norm{x_0} + \norm{\epsilon} =: r,
\end{align}
where $r$ does not depend on $s$. With the growth bound on $v_\theta$,
\begin{align}
  \norm{\epsilon - x_0 - v_\theta(x_s, s)}
  \le \norm{\epsilon} + \norm{x_0} + A + B\norm{x_s}^k
  \le A + r + B r^k.
\end{align}
Squaring gives, for every $s \in [0,1]$,
\begin{align}
  g(s; x_0, \epsilon) \le G(x_0, \epsilon) := \left(A + r + B r^k\right)^2.
\end{align}
The bound $G$ does not depend on $s$, and it is integrable. Expanding the square and the powers of $r = \norm{x_0} + \norm{\epsilon}$ writes $G$ as a finite sum of terms $c\, \norm{x_0}^i \norm{\epsilon}^j$ with $c \ge 0$. Since $x_0$ and $\epsilon$ are independent, each term has expectation $c\, \av{\norm{x_0}^i}_{q(x_0)} \av{\norm{\epsilon}^j}_{q(\epsilon)}$, which is finite because all moments of $q(x_0)$ and of the Gaussian $q(\epsilon)$ are finite.

\textbf{Limit.} Let $s_k \to s$ in $[0,1]$. To conclude $\ell(s_k) \to \ell(s)$ we must exchange the limit with the expectation, which continuity of the integrand alone does not allow, because $\epsilon$ is unbounded. The dominated convergence theorem~\citep[Theorem~1.34]{rudin1987real} allows it under two conditions: the integrands $g(s_k; x_0, \epsilon)$ converge for every $(x_0, \epsilon)$, and a single function with finite expectation bounds all of them, which prevents probability mass from escaping to infinity. Both hold here. Since $g(\cdot\,; x_0, \epsilon)$ is continuous for every fixed $(x_0, \epsilon)$, as shown above, $g(s_k; x_0, \epsilon) \to g(s; x_0, \epsilon)$ for every $(x_0, \epsilon)$, and $0 \le g(s_k; x_0, \epsilon) \le G(x_0, \epsilon)$ for every $k$, with $\av{G}_{q(x_0)q(\epsilon)} < \infty$. Hence $\ell(s_k) \to \ell(s)$, so $\ell$ is continuous on the compact interval $[0,1]$, and therefore bounded.
\end{proof}

\end{document}